# Fingerprint: DWT, SVD Based Enhancement and Significant Contrast for Ridges and Valleys Using Fuzzy Measures

D.Bennet and Dr. S. Arumuga Perumal

**Abstract**— The performance of the Fingerprint recognition system will be more accurate with respect of enhancement for the fingerprint images. In this paper we develop a novel method for Fingerprint image contrast enhancement technique based on the discrete wavelet transform (DWT) and singular value decomposition (SVD) has been proposed. This technique is compared with conventional image equalization techniques such as standard general histogram equalization and local histogram equalization. An automatic histogram threshold approach based on a fuzziness measure is presented. Then, using an index of fuzziness, a similarity process is started to find the threshold point. A significant contrast between ridges and valleys of the best, medium and poor finger image features to extract from finger images and get maximum recognition rate using fuzzy measures. The experimental results show the recognition of superiority of the proposed method to get maximum performance up gradation to the implementation of this approach.

——————————  ◆  ——————————

## 1 INTRODUCTION

Fingerprint based identification has been one of the most successful biometric techniques used for personal identification. Each individual has unique for fingerprints. A fingerprint is the pattern of ridges and valleys and singular point [2] on the finger tip. In different approaches to extract the feature sets [1] from the finger images and also some basic processing algorithms to apply for higher recognition rate. To improve the performance of finger image comes from its contrast. Contrast enhancement is frequently referred to as one of the most important issues in image processing. Contrast is created by the difference in luminance reflected from two adjacent surfaces. In visual perception, contrast is determined by the difference in the color and brightness of an object with other objects. Our visual system is more sensitive to contrast than absolute luminance; therefore, we can perceive the considerable changes in illumination conditions. If the contrast of an image is highly concentrated on a specific range, the information may be lost in those areas which are excessively and uniformly concentrated.

- *D. Bennet, Research Scholar, Dept. of Computer Science and Engineering, M.S. University, Tirunelveli, Tamil Nadu, India.,*
- *Dr. S. Arumuga Perumal, Head & Associate Professor, Dept. of Computer Science, S.T. Hindu College, Nagercoil-2., India.*

To improve the contrast enhancement several techniques are used base of general histogram equalization and local histogram equalization. We have proposed a new method for fingerprint image equalization is based on the SVD of an LL subband image obtained by DWT. DWT is used to separate the input low contrast finger image into different frequency subbands, where the LL subband concentrates the illumination information. Histograms of images with two distinct regions are formed by two peaks separated by a deep valley called bimodal histograms. In such cases, the threshold value must be located on the valley region. In case of thresholding techniques might perform poorly. Fuzzy set theory provides a new tool to deal with multimodal histograms. It can incorporate human perception and linguistic concepts such as similarity, and has been successfully applied to image thresholding the resultant image will be sharper with good contrast.

## 2 PROPOSED IMAGE CONTRAST ENHANCEMENTS

There are three significant parts for this proposed method. The first one is the use of SVD. As it was mentioned, the singular value matrix obtained by SVD [4] contains the illumination information. Therefore, changing the singular values will directly affect the illumination of the image; hence, the other information in the image will not be changed. The second important aspect of this work is the application of DWT. The illumination information is embedded in the LL subband. The edges are concentrated in the other subbands (i.e., LH, HL, and HH). Hence, separating the high-frequency subbands and applying the illumination enhancement in the LL subband only will protect the edge information from possible degradation. After the reconstruction the final image by using DWT. The resultant image is applying the fuzzy logic concepts; the problems involved in finding the minimum of a criterion function are avoided. Similarity between gray levels is the key to find an optimal threshold.

SVD was used to deal with an illumination problem. The method uses the ratio of the largest singular value of the generated normalized matrix, with mean zero and variance of one, over a normalized image which can be calculated the formula

$$\xi = \frac{\max\_\Sigma N(\mu=0, var=1)}{max(\Sigma A)} \qquad (1)$$

Where $\Sigma N$ ($\mu$=0, var=1) is the singular value matrix of the synthetic intensity matrix.

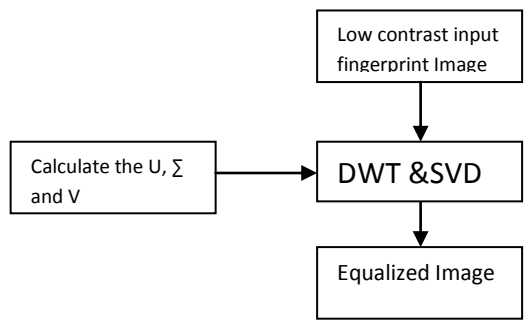

Fig. 1. Flow chart for proposed method

**2.1. Fuzzy Set Theory**

Fuzzy set theory assigns a membership degree to all elements among the universe of discourse according to their potential to fit in some class. The membership degree can be expressed by a mathematical function that assigns, to each element in the set, a membership degree between 0 and 1. A fuzzy set in is defined as

$$A = \{(xx_i, \mu_A(xi) | x_i \in X\} \qquad (2)$$

However, in images with low contrast, the method performs poorly due to the fact that one of the initial regions contains a low number of pixels. So, previous histogram equalization is carried out in images with low contrast aiming to provide an image with significant contrast [3]. For discrete values the cumulative distribution function is

$$T(x_i) = \sum_{k=0}^{i} p(x_k) = \sum_{k=0}^{i} \frac{n(x_k)}{MN} \qquad (3)$$

Thus, a processed image is obtained by mapping each pixel with $x_i$ level in the input image into a corresponding pixel with level $s_i = T(x_i)$ in the output finger image

## 3 FINGERPRINT IMAGE ENHANCEMENT

### 3.1 Existing Method

Otsu's method is used for global image threshold to get the poor result for different quality finger images. In a fingerprint image processing Otsu's method is used to automatically perform histogram shape-based image thresholding. The algorithm assumes that the image to be thresholded contains two classes of pixels (e.g. ridges and normalized and regarded as a probability valleys) then calculates the optimum threshold separating those two Measures s so that their combined spread (intra-class variance) is minimal. Let the pixels of given picture be represented in L gray levels [1, 2,…L]. The number of pixels with level I is denoted by ni and the total number of pixels by N = n1 + n2 + …+ nL. In order to simplify the discussion, the gray-level histogram is distribution:

$$p_i = n_i/N, \quad p_i \geq 0, \quad \sum_{i=0}^{L} p_i = 1 \qquad (4)$$

Now suppose that we dichotomize the pixel into two classes C0 and C1 (background and objects) by a threshold at level k: C0 denotes pixels with levels [0, k] and C 1denotes pixels with levels [k+1… L].

Then the probabilities of class occurrence and the class mean levels, respectively, are given by1

### 3.2 Proposed Method

Contrast enhancement techniques are used and implemented in fingerprint image processing and improve the recognition rate using SVD, DWT and Fuzzy Measures to get goodness of result. The different quality finger images are implemented with MATLAB code.

## 4 Experimental Results

In this experiment to compare and compute the different quality fingerprint images commonly best, medium and low quality images shown below

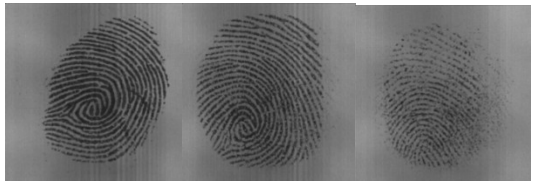

(a)  Best       (b) Medium        (c) Poor

Fig 2.Fingerprint images for Processing

In first case the Otsu Method is applied and the above finger image is enhanced to get the result show bellow fig.3.4 and 5.

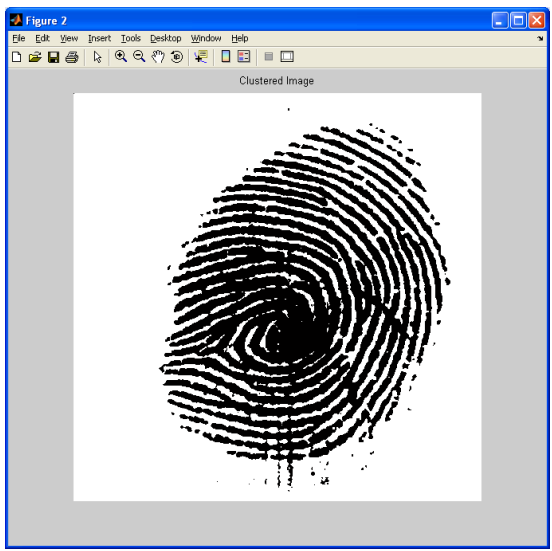

Fig.3. After Otsu's method applied best quality

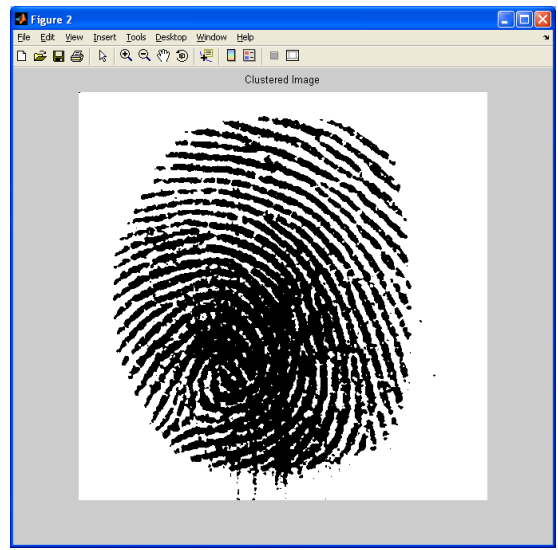

Fig.4., After Otsu's method applied medium quality

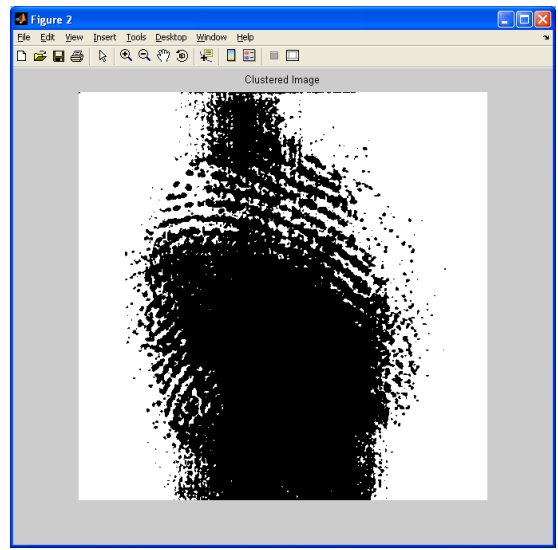

Fig.5., After Otsu's method applied Poor quality

In the second case the same fingerprint images are processed and compute the proposed method. This method to calculate the turned parameter based threshold   and calculate the SVD to adjust the approximation and directly extract the original parameters from DWT then the two extracted features are reconstructed. The feature set to turned the threshold values to improve the contrast for enhance the fingerprint image give more accuracy to findings the features in case of matching. To calculate the FAR/FRR and to reduce the noise also improved the recognition rate. To separate the ridges and valleys of the different quality fingerprint image get best enhancement the performance of the above method are shown in the fig. 6, 7 and 8 .

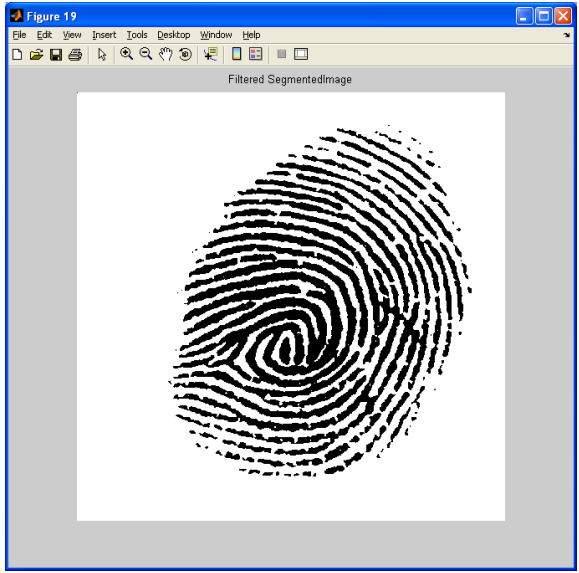

Fig.6.,After the Proposed method applied best quality

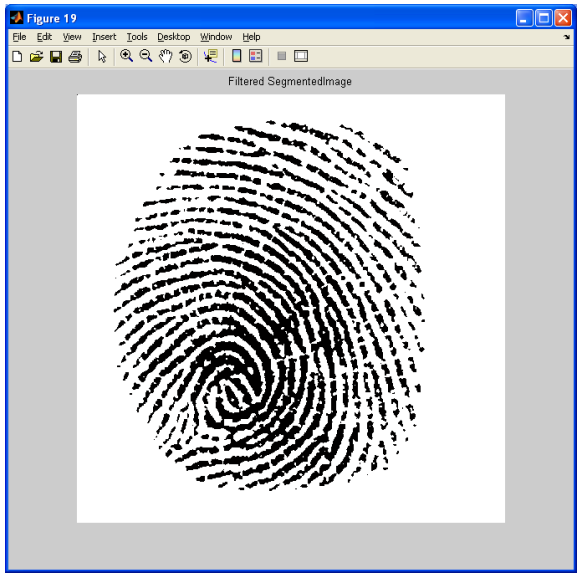

Fig.7.,After the Proposed method applied medium quality

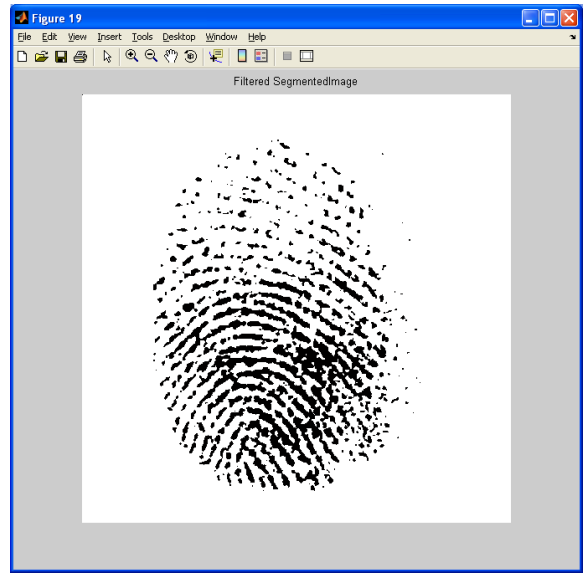

Fig.8.,After the Proposed method applied best quality

## 5. CONCLUSION

The above methods are implemented and to compare the outputs, the performance of the SVD, DWT and Fuzzy Measures method is more accurate contrast enhancement compare to Otsu method. After results analysis we can conclude that the proposed approach presents a higher performance for a large number of tested images.

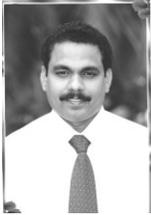

**Mr. D. Bennet**. Scholar, Manonmaniam Sundranar University, Tirunelveli. He is working as Asst. Professorr in South Travancore Hindu College for the last 11 years. He has completed his M.C.A., in Manonmaniam Sundranar University. M.Phil., Computer Science degree in Alagappa university, Karaikudi. He is a counselor in IGNOU and also staff in charge and examiner in Maduri Kamaraj University, Manonmaniam Sundranar University and Alagappa University. He has attended number of National and International seminars, conferences, Workshops and also presented papers. His area of research is Network Security using Biometrics. He has also deep knowledge in Digital Image Processing, Network Security, Multimedia and Data Mining.

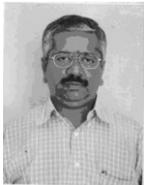

**Dr.S.Arumuga perumal** is working as a Associate Professor and Head of the Department of Computer science in South Travancore Hindu college for the last 20 years. He has completed his M.S(Software systems) in BITS, Pilani, Rajasthan; M.Phil Computer Science degree in Alagappa niversity, Karaikudi. And he did his Ph.D(Software Systems-Computer science) in Manonmanium Sundaranar University. He is a senior member of Computer society of India. He is a fellow of IEEE, IETE. He is involved in various academic activities. He has attended number of national and international seminars, conferences and presented number of papers. He has also published number of research articles in national and international journals. He is a member of curriculum development in universities and autonomous colleges. His area of research is Digital Image compression, Data mining and Biometrics.